\documentclass[final]{cvpr}
\usepackage{times}
\usepackage{epsfig}
\usepackage{graphicx}
\usepackage{amsmath}
\usepackage{amssymb}
\usepackage{upgreek}
\usepackage{booktabs}
\usepackage{diagbox}
\usepackage{multirow}

\makeatletter
\@namedef{ver@everyshi.sty}{}
\makeatother
\usepackage{tikz}
\usepackage{pgfplots}
\usepackage{caption}
\usepackage{graphicx}
\usepackage{pifont}
\newcommand{\cmark}{\ding{51}}%

\newlength\savewidth\newcommand\shline{\noalign{\global\savewidth\arrayrulewidth
  \global\arrayrulewidth 1pt}\hline\noalign{\global\arrayrulewidth\savewidth}}

\makeatletter\renewcommand\paragraph{\@startsection{paragraph}{4}{\z@}
  {.5em \@plus1ex \@minus.2ex}{-.5em}{\normalfont\normalsize\bfseries}}\makeatother
  
\usepackage[pagebackref=true,breaklinks=true,colorlinks,bookmarks=false]{hyperref}

\def\cvprPaperID{446} 

\def\cvprPaperID{446} 
\def\confYear{CVPR 2021}

\begin{document}

\title{Semi-Supervised Semantic Segmentation
with Cross Pseudo Supervision}  

\author{
Xiaokang Chen$^1$$\thanks{This work was done when Xiaokang Chen was an intern at Microsoft Research, Beijing, P.R. China}$ \quad
Yuhui Yuan$^2$ \quad
Gang Zeng$^1$ \quad
Jingdong Wang$^2$ \\[1.2mm]
$^1$Key Laboratory of Machine Perception (MOE), Peking University ~~\quad
$^2$Microsoft Research Asia
}

\maketitle

\begin{abstract}
In this paper,
we study the semi-supervised semantic segmentation problem
via exploring both labeled data and extra unlabeled data.
We propose a novel consistency regularization approach,
called cross pseudo supervision (CPS).
Our approach 
imposes the consistency on
two segmentation networks
perturbed with different initialization
for the same input image.
The pseudo one-hot label map,
output from one perturbed segmentation network,
is used to supervise the other segmentation network
with the standard cross-entropy loss,
and vice versa.
The CPS consistency has two roles: encourage high similarity
between the predictions of two perturbed networks
for the same input image,
and expand training data
by using the unlabeled data with pseudo labels.
Experiment results show that
our approach achieves the state-of-the-art semi-supervised segmentation performance
on Cityscapes and PASCAL VOC 2012. Code is available at \href{https://git.io/CPS}{https://git.io/CPS}.
\end{abstract}

\section{Introduction}
Image semantic segmentation 
is a fundamental recognition task
in computer vision.
The semantic segmentation training data 
requires pixel-level manual labeling,
which is much more expensive 
compared to the other vision tasks,
such as image classification and object detection.
This makes semi-supervised segmentation an important problem
to learn segmentation models
by using the labeled data as well as 
the additional unlabeled data.

Consistency regularization is widely studied in semi-supervised semantic segmentation. 
It enforces the consistency of the predictions with various perturbations,
e.g., input perturbation by augmenting input images~\cite{french2020semi,KimJP20},
feature perturbation~\cite{ouali2020semi},
and network perturbation~\cite{ke2019dual}.
Self-training is also studied for semi-supervised segmentation~\cite{ChenLCCCZAS20, ZophGLCLCL20, ZhuZWZHZMLS20, FengZCTSM20, hung2018adversarial, MittalIB19}.
It incorporates pseudo segmentation maps on the unlabeled images
obtained from the segmentation model trained on the labeled images
to expand the training data,
and retrains the segmentation model.

We present a novel and simple consistency regularization approach
with network perturbation, called cross pseudo supervision.
The proposed approach feeds the labeled and unlabeled images
into two segmentation networks 
that share the same structure and are initialized differently.
The outputs of the two networks on the labeled data are supervised separately
by the corresponding ground-truth segmentation map.
Our main point lies in the cross pseudo supervision that
enforces the consistency between the two segmentation networks.
Each segmentation network for an input image
estimates a segmentation result,
called pseudo segmentation map.
The pseudo segmentation map is used as an additional signal
to supervise
the other segmentation network.

The benefits from the cross pseudo supervision scheme  
lie in two-fold.
On the one hand,
like previous consistency regularization,
the proposed approach encourages that the predictions across differently initialized networks
for the same input image are consistent
and that the prediction decision boundary lies in low-density regions. 
On the other hand,
during the later optimization stage
the pseudo segmentation becomes stable
and more accurate than 
the result from normal supervised training
only on the labeled data.
The pseudo labeled data behaves like expanding the training data,
thus improving the segmentation network training quality.

Experimental results with various settings on two benchmarks,
Cityscapes and PASCAL VOC $2012$,
show that the proposed cross pseudo supervision approach is superior to existing consistency schemes for semi-supervised segmentation.
Our approach achieves the state-of-the-art semi-supervised segmentation performance on both benchmarks.

\section{Related work}

\noindent\textbf{Semantic segmentation.}
Modern deep learning methods for semantic segmentation
are mostly based on 
fully-convolutional network (FCN)~\cite{long2015fully}.
The subsequent developments 
studies the models
from three main aspects:
resolution, context, 
and edge.
The works on resolution enlargement
include mediating the spatial loss
caused in the classification network,
e.g., using the encoder-decoder scheme~\cite{chen2018encoder}
or dilated convolutions~\cite{yu2015multi,chen2018deeplab},
and maintaining high resolution, such as HRNet~\cite{wang2020deep,sun2019deep}.

The works on exploiting contexts
include spatial context, e.g., PSPNet~\cite{zhao2017pyramid} and ASPP~\cite{chen2018deeplab},
object context~\cite{yuan2018ocnet,yuan2019object},
and application of self-attention~\cite{vaswani2017attention}.
Improving the segmentation quality
on the edge areas include Gated-SCNN~\cite{takikawa2019gated},
PointRend~\cite{kirillov2020pointrend}, and SegFix~\cite{yuan2020segfix}.
In this paper,
we focus on how to use the unlabeled data,
conduct experiments mainly using DeepLabv$3$+
and also report the results on HRNet.

\vspace{.1cm}
\noindent\textbf{Semi-supervised semantic segmentation.}
Manual pixel-level annotations
for semantic segmentation is very time-consuming and costly.
It is valuable to explore the available unlabeled images
to help learn segmentation models.

Consistency regularization is widely studied
for semi-supervised segmentation.
It enforces the consistency of the predictions/intermediate features with various perturbations.
Input perturbation methods~\cite{french2020semi,KimJP20}
augment the input images randomly
and impose the consistency constraint
between the predictions of augmented images,
so that the decision function lies 
in the low-density region.

Feature perturbation presents a feature perturbation scheme by using multiple decoders
and enforces the consistency between the outputs
of the decoders~\cite{ouali2020semi}.
The approach GCT~\cite{ke2020guided} further performs 
network perturbation
by using two segmentation networks 
with the same structure but initialized differently
and enforces the consistency between the predictions
of the perturbed networks.
Our approach differs from GCT
and enforces the consistency by using the pseudo segmentation maps with an additional benefit
like expanding the training data.

Other than enforcing the consistency 
between various perturbations for one image,
the GAN-based approach~\cite{MittalIB19}
enforce the consistency
between the statistical features
of the ground-truth segmentation maps for labeled data
and the predicted segmentation maps on unlabeled data.
The statistical features
are extracted from a discriminator network
that is learned to
distinguish ground-truth segmentation
and predicted segmentation.

Self-training, a.k.a., self-learning,
self-labeling,
or decision-directed learning,
is initially developed
for using unlabeled data in classification~\cite{Scudder65a, Fralick67, Agrawala70, CSZ2006, lee2013pseudo}.
Recently it is applied for semi-supervised segmentation~\cite{ChenLCCCZAS20, ZophGLCLCL20, ZhuZWZHZMLS20, FengZCTSM20, hung2018adversarial, MittalIB19,ibrahim2020semi,mendel2020semi}.
It incorporates pseudo segmentation maps on unlabeled data
obtained from the segmentation model previously trained on labeled data
for retraining the segmentation model.
The process can be iterated several times.
Various schemes are introduced 
on how to decide the pseudo segmentation maps.
For example,
the GAN-based methods~\cite{hung2018adversarial, MittalIB19,souly2017semi},
use the discriminator learned for distinguishing
the predictions and the ground-truth segmentation
to select high-confident segmentation predictions on unlabeled images
as pseudo segmentation.

PseudoSeg~\cite{zou2020pseudoseg},
concurrent to our work, 
also explores pseudo segmentation for semi-supervised segmentation.
There are at least two differences from our approach.
PseudoSeg follows the FixMatch scheme~\cite{SohnBLZCCKZR20}
via using the pseudo segmentation of a weakly-augmented image
to supervise the segmentation of a strongly-augmented image
based on a single segmentation network.
Our approach adopts two same and independently-initialized segmentation networks
with the same input image,
and uses the pseudo segmentation maps of each network
to supervise the other network.
On the other hand,
our approach performs back propagation
on both the two segmentation networks, while PseudoSeg only performs back propagation
for the strongly-augmented image.

\vspace{.1cm}
\noindent\textbf{Semi-supervised classification.}
Semi-supervised classification was widely studied
in the first decade of this century~\cite{CSZ2006}.
Most solutions are based on the assumptions, such as smoothness, consistency,
low-density, or clustered.
Intuitively, neighboring data have a high probability of belonging to the same class,
or the decision boundary should lie in low-density regions.

Deep learning methods 
impose the consistency 
over perturbed inputs or augmented images
encouraging the model 
to produce the similar output/distributions
for the perturbed inputs,
such as temporal ensembling~\cite{LaineA17}
and its extension mean teacher~\cite{tarvainen2017mean}.
Dual student~\cite{ke2019dual} makes modifications
by jointly learning
two classification networks that initialized differently
with complex consistency on the predictions
and different image augmentations.

Other development includes estimating labels
for unlabeled data,
e.g., MixMatch~\cite{BerthelotCGPOR19} combining the estimations
on multiple augmentations,
FixMatch~\cite{SohnBLZCCKZR20} using pseudo labels on weak augmentation
to supervise the labeling on strong augmentation.

\begin{figure*}[t]
\centering
\footnotesize
(a)~~
\includegraphics[width=.45\linewidth]{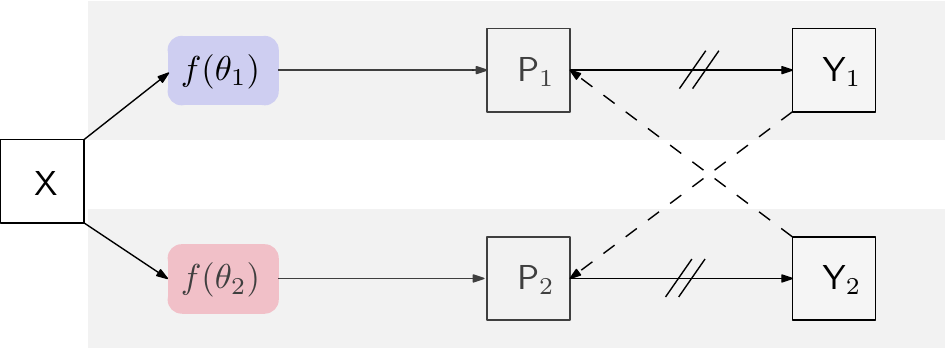}~~~~
(b)~~\includegraphics[width=.45\linewidth]{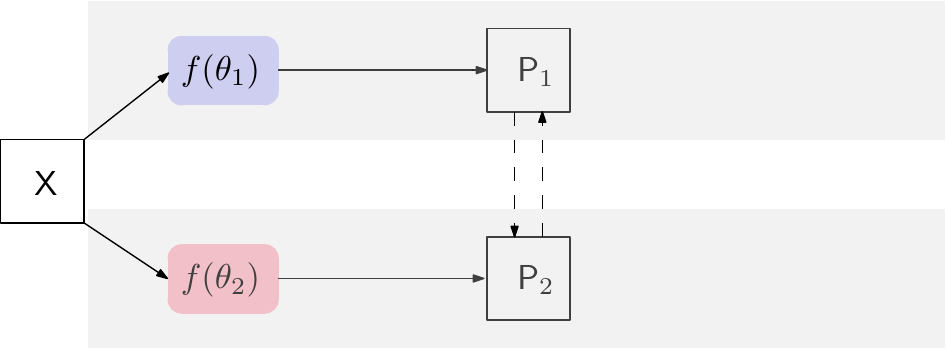} \\
\vspace{.1cm}
(c)~~~\includegraphics[width=.45\linewidth]{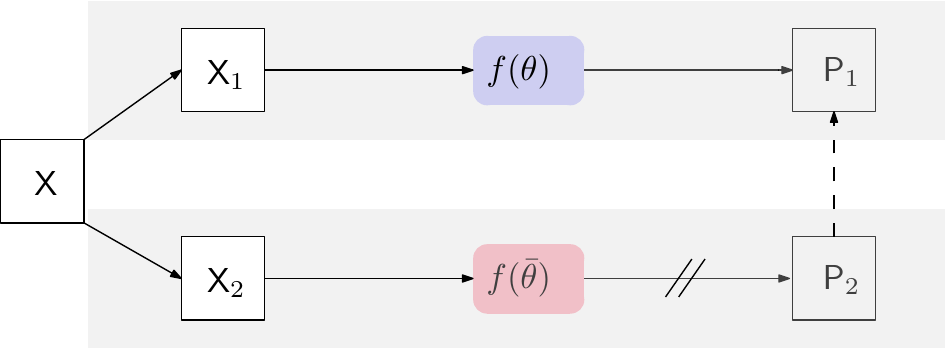}~~~~
(d)~~\includegraphics[width=.45\linewidth]{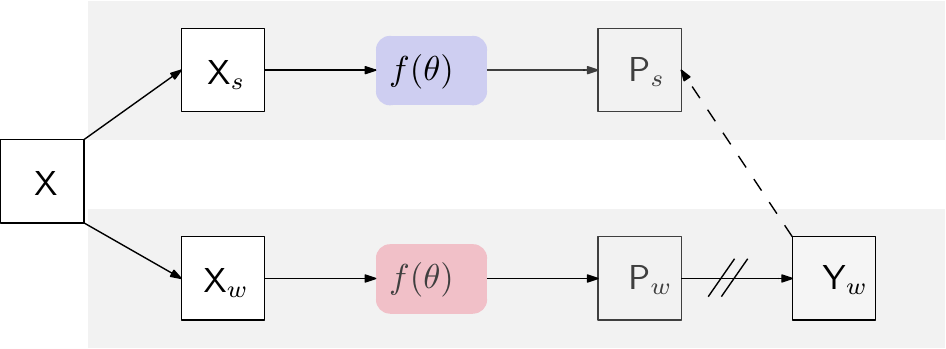}
\captionsetup{font={small}}
\caption{Illustrating the architectures for 
(a) our approach cross pseudo supervision,
(b) cross confidence consistency (e.g., a component of GCT~\cite{ke2020guided}),
(c) mean teacher (used in CutMix-Seg~\cite{french2020semi}),
and (d) PseudoSeg~\cite{zou2020pseudoseg} structure (similar to FixMatch~\cite{SohnBLZCCKZR20}). `$\rightarrow$' means forward operation and `$\dashrightarrow$' means loss supervision.
`$//$' on `$\rightarrow$' means stop-gradient.
More details are illustrated in the approach section.
}
\label{fig:structure}
\vspace{-4mm}
\end{figure*}

\section{Approach}
Given a set $\mathcal{D}^l$ of $N$ labeled images,
and a set $\mathcal{D}^u$ of $M$ unlabeled images,
the semi-supervised semantic segmentation task
aims to learn a segmentation network
by exploring both the labeled and unlabeled images.

\vspace{.1cm}
\noindent\textbf{Cross pseudo supervision.}
The proposed approach consists of 
two parallel segmentation networks:
\begin{align}
\mathsf{P}_1 & = f(\mathsf{X}; \boldsymbol{\uptheta}_1),\\
\mathsf{P}_2 & = f(\mathsf{X}; \boldsymbol{\uptheta}_2).
\end{align}
The two networks have the same structure
and their weights, i.e., $\boldsymbol{\uptheta}_1$ and $\boldsymbol{\uptheta}_2$ , are initialized differently.
The inputs $\mathsf{X}$ are with the same augmentation,
and $\mathsf{P}_1$ ($\mathsf{P}_2$) 
is the segmentation confidence map,
which is the network output after softmax normalization.
The proposed approach is logically illustrated as below\footnote{We use $f(\boldsymbol{\uptheta})$
to represent $f(\mathsf{X}; \boldsymbol{\uptheta})$
by dropping $\mathsf{X}$ for convenience.}:
\begin{align}
\mathsf{X} \rightarrow \mathsf{X} & \rightarrow f(\boldsymbol{\uptheta}_1) \rightarrow \mathsf{P}_1 \rightarrow \mathsf{Y}_1 \notag \\
& \searrow f(\boldsymbol{\uptheta}_2) \rightarrow \mathsf{P}_2 \rightarrow \mathsf{Y}_2.
\end{align}
Here $\mathsf{Y}_1$ ($\mathsf{Y}_2$)
is the predicted one-hot label map,
called pseudo segmentation map.
At each position $i$,
the label vector $\mathbf{y}_{1i}$ 
($\mathbf{y}_{2i}$ )
is a one-hot vector
computed from the corresponding confidence vector
$\mathbf{p}_{1i}$ 
($\mathbf{p}_{2i}$ ).
The complete version of our method is illustrated in Figure~\ref{fig:structure} (a)
and we have not included the loss supervision
in the above equations.

The training objective contains two losses:
supervision loss $\mathcal{L}_s$
and cross pseudo supervision loss $\mathcal{L}_{cps}$.
The supervision loss $\mathcal{L}_s$ 
is formulated using
the standard pixel-wise cross-entropy loss
on the labeled images
over the two parallel segmentation networks:
\begin{align}
    \mathcal{L}_s = \frac{1}{|\mathcal{D}^l|}
    \sum_{\mathsf{X} \in \mathcal{D}^l}
    \frac{1}{W \times H} \sum_{i = 0}^{W\times H} 
    (\ell_{ce}(\mathbf{p}_{1i}, \mathbf{y}^*_{1i}) \notag\\
    + \ell_{ce}(\mathbf{p}_{2i}, \mathbf{y}^*_{2i})),
\end{align}
where $\ell_{ce}$ is the cross-entropy loss function and $\mathbf{y}^*_{1i} (\mathbf{y}^*_{2i})$ is the ground truth. $W$ and $H$ represent the width and height of the input image.

The cross pseudo supervision loss
is bidirectional:
One is from $f(\uptheta_1)$ to $f(\uptheta_2)$.
We use the pixel-wise one-hot label map
$\mathsf{Y}_1$
output from one network $f(\uptheta_1)$
to supervise the pixel-wise confidence map $\mathsf{P}_2$
of the other network $f(\uptheta_2)$,
and the other one
is from $f(\uptheta_2)$ to $f(\uptheta_1)$.
The cross pseudo supervision loss
on the unlabeled data is written as
\begin{align}
    \mathcal{L}_{cps}^u = \frac{1}{|\mathcal{D}^u|}
    \sum_{\mathsf{X} \in \mathcal{D}^u}
    \frac{1}{W \times H} \sum_{i = 0}^{W\times H}
    (\ell_{ce}(\mathbf{p}_{1i}, \mathbf{y}_{2i}) \notag\\
    + \ell_{ce}(\mathbf{p}_{2i}, \mathbf{y}_{1i})).
\end{align}

We also define the cross pseudo supervision loss $\mathcal{L}_{cps}^l$ on the labeled data
in the same way.
The whole cross pseudo supervision loss
is the combination of the losses on
both the labeled and unlabeled data:
$\mathcal{L}_{cps} = \mathcal{L}_{cps}^l + \mathcal{L}_{cps}^u$.

The whole training objective is written as:
\begin{align}
\label{eq:whole-loss}
\mathcal{L} = \mathcal{L}_s
+ \lambda \mathcal{L}_{cps},
\end{align}
where $\lambda$
is the trade-off weight.

\vspace{.1cm}
\noindent\textbf{Incorporation with 
the CutMix augmentation.} \label{ourmethod_cutmix}
The CutMix augmentation scheme~\cite{yun2019cutmix}
is applied to the mean teacher framework for semi-supervised segmentation~\cite{french2020semi}.
We also apply the CutMix augmentation in our approach.
We input the CutMixed image into the two networks $f(\boldsymbol{\uptheta}_1)$ and $f(\boldsymbol{\uptheta}_2)$.
We use the way similar to~\cite{french2020semi}
to generate pseudo segmentation maps from the two networks:
input two source images (that are used to generate the CutMix images)
into each segmentation network
and mix the two pseudo segmentation maps
as the supervision of the other segmentation network.

\section{Discussions}
We discuss the relations of our method
with several related works as following.

\vspace{.1cm}
\noindent\textbf{Cross probability consistency.} 
An optional consistency across the two perturbed networks
is cross probability consistency:
the probability vectors 
(from pixel-wise confidence maps)
should be similar
(illustrated in Figure~\ref{fig:structure} (b)).
The loss function is written as:
\begin{align}
    \mathcal{L}_{cpc} = \frac{1}{|\mathcal{D}|}
    \sum_{\mathsf{X} \in \mathcal{D}}
    \frac{1}{W \times  H} \sum_{i = 0}^{W \times H}
    (\ell_{2}(\mathbf{p}_{1i}, \mathbf{p}_{2i}) \notag \\
    + \ell_{2}(\mathbf{p}_{2i}, \mathbf{p}_{1i})).
\end{align}

Here an example loss $\ell_2(\mathbf{p}_{1i}, \mathbf{p}_{2i}) = \|\mathbf{p}_{1i} - \mathbf{p}_{2i}\|_2^2$ is used to impose the consistency.
Other losses, such as KL-divergence, 
and the consistency over the intermediate features can also be used.
We use $\mathcal{D}$ to represent the union of labeled set $\mathcal{D}^l$ and unlabeled set $\mathcal{D}^u$.

Similar to the feature/probability consistency,
the proposed cross pseudo supervision consistency
also expects the consistency between the two perturbed segmentation networks. 
In particular, our approach in some sense augments
the training data by exploring the unlabeled data with pseudo labels.
The empirical results shown in Table~\ref{tab:abla-of-losses}
indicates that
cross pseudo supervision outperforms
cross probability consistency.

\vspace{.1cm}
\noindent\textbf{Mean teacher.}
Mean teacher~\cite{tarvainen2017mean} is initially developed for semi-supervised classification
and recently applied for semi-supervised segmentation, e.g., in CutMix-Seg~\cite{french2020semi}.
The unlabeled image with different augmentations is fed into two networks
with the same structure: 
one is student $f(\uptheta)$,
and the other one is mean teacher
$f(\bar{\uptheta})$
with the parameter $\bar{\uptheta}$
being the moving average of the student network parameter $\uptheta$:
\begin{align}
\mathsf{X} & \rightarrow \mathsf{X}_1  \rightarrow f(\boldsymbol{\uptheta}) \rightarrow \mathsf{P}_1  \notag \\
& \searrow \mathsf{X}_2 \rightarrow f(\bar{\boldsymbol{\uptheta}}) \nrightarrow \mathsf{P}_2. 
\end{align}

We use $\mathsf{X}_1$ and $\mathsf{X}_2$ to represent the differently augmented version
of $\mathsf{X}$.
The consistency regularization aims
to align the probability map $\mathsf{P}_1$ of $\mathsf{X}_1$ predicted by the student network 
to the probability map $\mathsf{P}_2$ of $\mathsf{X}_2$ predicted by the teacher network.
During the training,
we supervise $\mathsf{P}_1$ with $\mathsf{P}_2$ and
apply no back propagation for the teacher network.
We use $\nrightarrow$ to represent ``no back propagation'' in the following illustration.
we have not included the loss supervision
in the above equations and illustrate the complete version in Figure~\ref{fig:structure} (c).
The results in Table~\ref{tab:sota-VOC-subset} and Table~\ref{tab:sota-City-subset} show that our approach is superior to the
mean teacher approach.

\vspace{.1cm}
\noindent\textbf{Single-network
pseudo  supervision.}
We consider a downgraded version of our approach,
single-network pseudo supervision,
where the two networks are the same:
\begin{align}
\mathsf{X} \rightarrow \mathsf{X} & \rightarrow f(\boldsymbol{\uptheta}) \rightarrow \mathsf{P}  \nwarrow \notag \\
& \searrow f(\boldsymbol{\uptheta}) \rightarrow \mathsf{P} \nrightarrow \mathsf{Y}.
\label{eqn:emselftraining}
\end{align}

The structure is similar to Figure~\ref{fig:structure} (d),
and the only difference is that
the inputs to two streams
are the same
rather than one weak augmentation
and one strong augmentation.
We use $\nwarrow$ from $\mathsf{Y}$ to $\mathsf{P}$ to
represent the loss supervision.

Empirical results show that
single-network
pseudo  supervision
performs poorly.
The main reason
is
that
supervision by the pseudo label from the same network tends to learn the network itself
to better approximate the pseudo labels
and thus the network might converge in the wrong direction.
In contrast,
supervision by the cross pseudo label from the other network,
which differs from the pseudo label
from the network itself
due to network perturbation,
is able to learn the network
with some probability away from the wrong direction. 
In other words,
the perturbation of pseudo label
between two networks
in some sense serves
as a regularizer,
free of over-fitting the wrong direction.

In addition, we study
the single-network pseudo supervision 
in a way like~\cite{french2020semi}
with the CutMix augmentation.
We input two source images
into a network $f(\boldsymbol{\uptheta})$
and mix the two pseudo segmentation maps
as a pseudo segmentation map of the CutMixed image,
which is used to supervise
the output of the CutMixed image
from the same network.
Back propagation of the pseudo supervision is only done for the CutMixed image.
The results show that our approach performs better (Table~\ref{tab:abla-single-network-sup}),
implying that network perturbation is helpful
though there is already perturbation from
the way with the CutMix augmentation
in~\cite{french2020semi}.

\vspace{.1cm}
\noindent\textbf{PseudoSeg.}
PseudoSeg~\cite{zou2020pseudoseg}, similar to FixMatch~\cite{SohnBLZCCKZR20},
applies weakly-augmented image $\mathsf{X}_w$ to generate pseudo segmentation map,
which is used to supervise the output
of strongly-augmented image $\mathsf{X}_s$ from the same network
with the same parameters.
$\mathsf{X}_w$ and $\mathsf{X}_s$ are based on the same input image $\mathsf{X}$.
PseudoSeg only conducts back propagation on the path that processes
the strongly-augmented image $\mathsf{X}_s$
(illustrated in Figure~\ref{fig:structure} (d)).
It is logically formed as:
\begin{align}
\mathsf{X} & \rightarrow \mathsf{X}_s  \rightarrow f(\boldsymbol{\uptheta}) \rightarrow \mathsf{P}_s \nwarrow \notag \\
& \searrow \mathsf{X}_w \rightarrow f(\boldsymbol{\uptheta}) \rightarrow \mathsf{P}_w \nrightarrow \mathsf{Y}_w. 
\end{align}

We use $\nwarrow$ from $\mathsf{Y}_w$ to $\mathsf{P}_s$ to
represent the loss supervision.
The above manner is similar to single-network pseudo supervision.
The difference is that the pseudo segmentation map is from weak augmentation 
and it supervises the training over strong augmentation.
We guess that besides the segmentation map based on weak augmentation is more accurate,
the other reason is same as our approach:
the pseudo segmentation map from weak augmentation
also introduces extra perturbation to the pseudo supervision.

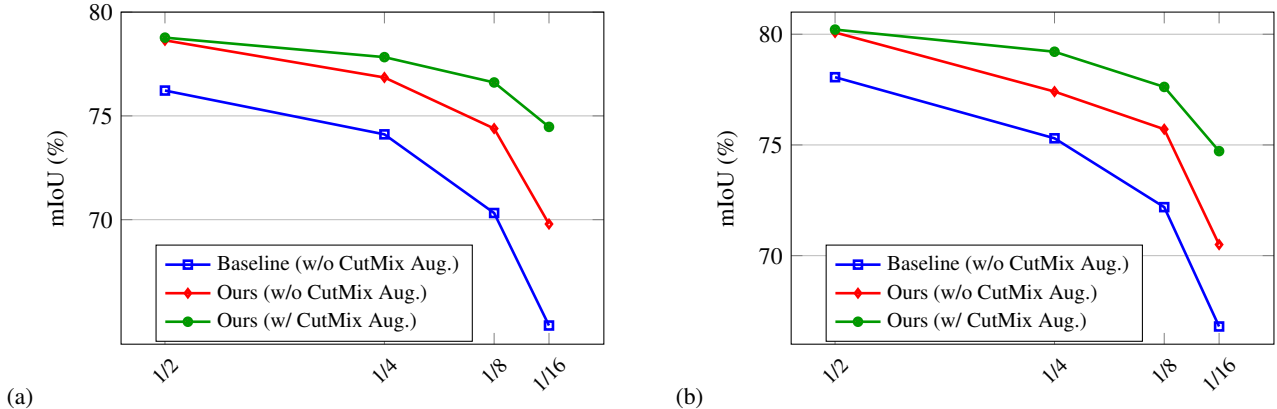
\begin{figure*}[t]
\centering
\small
(a)
\begin{tikzpicture}[scale=1]                 
\begin{axis}[legend columns=1, legend style={at={(0.4,0.3)},anchor=north,font=\footnotesize}, legend cell align={left},
y label style={at={(0.06,0.5)}},
ylabel={mIoU (\%)}, ymajorgrids=true, xtick={0.0625,0.125,0.25,0.5}, xticklabels={1/16, 1/8, 1/4, 1/2},
ytick={70, 75, 80},
yticklabels={70, 75, 80},
xticklabel style={rotate=45,font=\footnotesize},
x dir=reverse,
height=6cm,
width=8cm,
xmin=0, ymax=80, ymin=64]
\addplot+[sharp plot, line width=1pt, mark size=1.5pt, mark=square]
table
{
X Y
0.5 76.22
0.25 74.11
0.125 70.32
0.0625 64.90
};

\addplot+[sharp plot, line width=1pt, mark size=1.5pt, mark=diamond, color=red]
table
{
X Y
0.5 78.64
0.25 76.85
0.125 74.39
0.0625 69.79
};

\addplot+[sharp plot, line width=1pt, mark size=1.5pt, color=green!60!black]
table
{
X Y
0.5 78.77
0.25 77.83
0.125 76.61
0.0625 74.47
};

\addlegendentry{Baseline (w/o CutMix Aug.)}    
\addlegendentry{Ours (w/o CutMix Aug.)} 
\addlegendentry{Ours (w/ CutMix Aug.)} 
\end{axis}
\end{tikzpicture}
~~~~~~~~~~~(b)
\begin{tikzpicture}[scale=1]                 
\begin{axis}[legend columns=1, legend style={at={(0.4,0.3)},anchor=north,font=\footnotesize}, legend cell align={left},
y label style={at={(0.06,0.5)}},
ylabel={mIoU (\%)}, ymajorgrids=true, xtick={0.0625,0.125,0.25,0.5}, xticklabels={1/16, 1/8, 1/4, 1/2},
xticklabel style={rotate=45,font=\footnotesize},
x dir=reverse,
ytick={70, 75, 80},
yticklabels={70, 75, 80},
height=6cm,
width=8cm,
xmin=0, ymax=81, ymin=66]
\addplot+[sharp plot, line width=1pt, mark size=1.5pt, mark=square]
table
{
X Y
0.5 78.06
0.25 75.30
0.125 72.19
0.0625 66.80
};

\addplot+[sharp plot, line width=1pt, mark size=1.5pt, mark=diamond, color=red]
table
{
X Y
0.5 80.08
0.25 77.41
0.125 75.71
0.0625 70.50
};

\addplot+[sharp plot, line width=1pt, mark size=1.5pt, color=green!60!black]
table
{
X Y
0.5 80.21
0.25 79.21
0.125 77.62
0.0625 74.72
};

\addlegendentry{Baseline (w/o CutMix Aug.)}    
\addlegendentry{Ours (w/o CutMix Aug.)} 
\addlegendentry{Ours (w/ CutMix Aug.)} 
\end{axis}
\end{tikzpicture}

\captionsetup{font={small}}
\caption{\textbf{Improvements over 
the supervised baseline} on the Cityscapes val set
with (a) ResNet-$50$ and (b) ResNet-$101$.
}
\label{fig:vsbaseline}
\vspace{-5mm}
\end{figure*}

\section{Experiments}

\subsection{Setup}
\noindent\textbf{Datasets.} 
\emph{PASCAL VOC ${2012}$}~\cite{everingham2015pascal}
is a standard object-centric semantic segmentation dataset,
which consists of more than $13,000$ images with 
$20$ object classes and $1$ background class.
The standard training, validation and test sets consist of $1,464$,\quad$1,449$ and $1,456$ images respectively.
We follow the previous work to use the augmented set~\cite{hariharan2011sbd} ($10,582$ images)
as our full training set.

\emph{Cityscapes}~\cite{cordts2016cityscapes}
is mainly designed for urban scene understanding. The official split has $2,975$ images for training, $500$ for validation and $1,525$ for testing. Each image has a resolution of $2048 \times 1024$ and is fine-annotated with pixel-level labels of $19$ semantic classes.

We follow the partition protocols of Guided Collaborative Training (GCT)~\cite{ke2020guided} and divide the whole training set to two groups via
randomly sub-sampling $1/2$, $1/4$, $1/8$
and $1/16$ of the whole set as the labeled set
and regard the remaining images as the unlabeled set.

\vspace{.1cm}
\noindent\textbf{Evaluation.}
We evaluate the segmentation performance using mean Intersection-over-Union (mIoU) metric. For all partition protocols, we report results on the $1,456$ PASCAL VOC ${2012}$ \texttt{val} set
(or $500$ Cityscapes \texttt{val} set) via only
single scale testing.
We only use one network in our approach
to generate the results for evaluation.

\vspace{.1cm}
\noindent\textbf{Implementation details.}
We implement our method based on PyTorch framework.
We initialize the weights of two backbones
in the two segmentation networks
with the same weights pre-trained on ImageNet
and the weights of two segmentation heads (of DeepLabv$3$+) randomly.
We adopt mini-batch SGD with momentum to train our model
with Sync-BN~\cite{ioffe2015batch}.
The momentum is fixed as $0.9$ and the weight decay is set to $0.0005$.
We employ a poly learning rate policy where the initial learning rate is multiplied by $(1-\frac{iter}{max\_iter})^{0.9}$.

For the supervised baseline trained on the full training set, we use random horizontal flipping and multi-scale as data augmentation if not specified. We train PASCAL VOC $2012$ for $60$ epochs with base learning rate set to $0.01$, and Cityscapes for $240$ epochs with base learning rate set to 0.04. OHEM loss is used on Cityscapes.

\subsection{Results}

\noindent\textbf{Improvements over baselines.}
We illustrate the improvements of our method compared with
the supervised baseline under all partition protocols in Figure~\ref{fig:vsbaseline}.
All the methods are based on DeepLabv$3$+ with ResNet-$50$ or ResNet-$101$.

Figure~\ref{fig:vsbaseline} (a)
shows our method consistently
outperforms the supervised baseline on Cityscapes with ResNet-$50$.
Specifically, the improvements of our method w/o CutMix augmentation
over the baseline method w/o CutMix augmentation are
$4.89\%$, $4.07\%$, $2.74\%$, and $2.42\%$
under $1/16$, $1/8$, $1/4$, and $1/2$ partition protocols separately.
Figure~\ref{fig:vsbaseline} (b) shows the gains of
our method over the baseline method on Cityscapes with ResNet-$101$:
$3.70\%$, $3.52\%$, $2.11\%$, and $2.02\%$
under $1/16$, $1/8$, $1/4$, and $1/2$ partition protocols separately.

Figure~\ref{fig:vsbaseline} also shows the improvements brought by the CutMix augmentation.
We can see that CutMix brings more gains under the $1/16$ and $1/8$ partitions
than under the $1/4$ and $1/2$ partitions.
For example, on Cityscapes with ResNet-$101$, the extra gains brought by CutMix augmentation are
$4.22\%$, $1.91\%$, and $0.13\%$
under $1/16$, $1/8$, and $1/2$ partition protocols separately.

\renewcommand{\arraystretch}{1.5}
\begin{table*}[]
\centering\setlength{\tabcolsep}{6.5pt}
\captionsetup{font={small}}
\caption{\textbf{Comparison with state-of-the-arts}
on the PASCAL VOC $2012$ val set
under different partition protocols.
{\color{black}All the methods are based on DeepLabv$3$+.}
}
\vspace{-2mm}
\label{tab:sota-VOC-subset}
\footnotesize
\begin{tabular}{l|c|c|c|c|c|c|c|c}
\shline 
\multirow{2}{*}{Method} & \multicolumn{4}{c|}{ResNet-$50$} & \multicolumn{4}{c}{ResNet-$101$} \\\cline{2-9}
& $1/16$ ($662$) & $1/8$ ($1323$) & $1/4$ ($2646$) & $1/2$ ($5291$) & $1/16$ ($662$) & $1/8$ ($1323$) & $1/4$ ($2646$) & $1/2$ ($5291$) \\
\shline
{\color{black}MT}~\cite{tarvainen2017mean}   & $66.77$ & $70.78$ & $73.22$ & $75.41$ & $70.59$ & $73.20$ & $76.62$ & $77.61$ \\
CCT~\cite{ouali2020semi} &  $65.22$ & $70.87$ & $73.43$ & $74.75$ & $67.94$ & $73.00$ & $76.17$ & $77.56$ \\
{\color{black}CutMix-Seg}~\cite{french2020semi} & $68.90$ & $70.70$ & $72.46$ & $74.49$ & \textbf{$72.56$} & $72.69$ & $74.25$ & $75.89$  \\
{\color{black} GCT}~\cite{ke2020guided} & $64.05$ & $70.47$ & $73.45$ & $75.20$ & $69.77$ & $73.30$ & $75.25$ & $77.14$\\ \hline
Ours (w/o CutMix Aug.) & $68.21$ & $ {73.20} $ & ${74.24} $ & $ 75.91 $ & $72.18$ & ${75.83} $ & $ {77.55} $ & $ {78.64} $ \\ 
Ours (w/ CutMix Aug.) & $\mathbf{71.98}$ & $ \mathbf{73.67} $ & $\mathbf{74.90} $ & $\mathbf{76.15} $ & $\bf{74.48}$ & $\bf{76.44}$ & $\bf{77.68}$ & $\bf{78.64}$ \\ \hline
\end{tabular}
\vspace{-1mm}
\end{table*}

\renewcommand{\arraystretch}{1.5}
\begin{table*}[]
\centering
\setlength{\tabcolsep}{7.5pt}
\captionsetup{font={small}}
\caption{\textbf{Comparison with state-of-the-arts}
on the Cityscapes val set under
different partition protocols.
{\color{black}All the methods are based on DeepLabv$3$+.}
}
\vspace{-2mm}
\centering
\label{tab:sota-City-subset}
\footnotesize
\begin{tabular}{l|c|c|c|c|c|c|c|c}
\shline 
\multirow{2}{*}{Method} & \multicolumn{4}{c|}{ResNet-$50$} & \multicolumn{4}{c}{ResNet-$101$} \\\cline{2-9}
& $1/16$ ($186$) & $1/8$ ($372$) & $1/4$ ($744$) & $1/2$ ($1488$) & $1/16$ ($186$) & $1/8$ ($372$) & $1/4$ ($744$) & $1/2$ ($1488$) \\
\shline
MT~\cite{tarvainen2017mean} &  $66.14$ & $72.03$ & $74.47$ & $77.43$ & $68.08$ & $73.71$ & $76.53$ & $78.59$ \\
CCT~\cite{ouali2020semi} & $66.35$ & $72.46$ & $75.68$ & $76.78$ & $69.64$ & $74.48$ & $76.35$ & $78.29$ \\
GCT~\cite{ke2020guided} & $65.81$ & $71.33$ & $75.30$ & $77.09$ & $66.90$ & $72.96$ & $76.45$ & $78.58$\\ \hline
Ours (w/o CutMix Aug.) & ${69.79}$ & ${74.39}$ & ${76.85}$ & $78.64$ & ${70.50}$ & ${75.71}$ & ${77.41}$ & ${80.08}$ \\ 
Ours (w/ CutMix  Aug.)  & $\mathbf{74.47}$ & $ \mathbf{76.61} $ & $\mathbf{77.83} $ & $ \mathbf{78.77} $ & $\mathbf{74.72}$ & $ \mathbf{77.62} $ & $\mathbf{79.21} $ & $ \mathbf{80.21} $ \\ 
\hline
\end{tabular}
\vspace{-2mm}
\end{table*}

\renewcommand{\arraystretch}{1.5}
\begin{table}[]
\centering
\setlength{\tabcolsep}{1pt}
\captionsetup{font={small}}
\caption{\textbf{Comparison with state of the arts}
on the Cityscapes val set under
different partition protocols
using HRNet-W$48$.
}
\vspace{-2mm}
\centering
\label{tab:sota-HRNet-subset}
\footnotesize
\begin{tabular}{l|c|c|c|c}
\shline 
\multirow{2}{*}{Method} & \multicolumn{4}{c}{Cityscapes} \\\cline{2-5} & $1/16$ ($186$) & $1/8$ ($372$) & $1/4$ ($744$) & $1/2$ ($1488$) \\
\shline
Base &  $66.90$ & $72.79$ & $75.23$ & $78.09$ \\
Ours (w/o CutMix Aug.) & $72.49$ & $76.32$ & $78.27$ & $80.02$ \\
Ours (w/ CutMix Aug.) & $\textbf{75.09}$ & $\textbf{77.92}$ & $\textbf{79.24}$ & $\textbf{80.67}$ \\
\hline
\end{tabular}
\vspace{-4mm}
\end{table}

\vspace{.1cm}
\noindent\textbf{Comparison with SOTA.}
We compare our method
with some recent semi-supervised segmentation methods including:
Meat-Teacher (MT)~\cite{tarvainen2017mean},
Cross-Consistency Training (CCT)~\cite{ouali2020semi},
Guided Collaborative Training (GCT)~\cite{ke2020guided},
and CutMix-Seg~\cite{french2020semi} under different partition protocols.
Specifically,
we adopt the official open-sourced implementation of CutMix-Seg.
For MT and GCT, we use implementations from~\cite{ke2020guided}.
We compare them using the same architecture and partition protocols for fairness.

\emph{PASCAL VOC $2012$}:
Table~\ref{tab:sota-VOC-subset} shows the comparison results on PASCAL VOC $2012$.
We can see that over all the partitions, with both ResNet-$50$
and ResNet-$101$,
our method w/o CutMix augmentation consistently outperforms 
the other methods
except CutMix-Seg
that uses the strong CutMix augmentation~\cite{yun2019cutmix}.

Our approach w/ CutMix augmentation performs the best
and sets new state-of-the-arts under
all partition protocols.
For example, our approach w/ CutMix augmentation
outperforms the CutMix-Seg by
$3.08\%$ and $1.92\%$ under $1/16$ partition protocol
with ResNet-$50$ and ResNet-$101$ separately.
The results imply
that our cross pseudo supervision scheme
is superior to mean teacher scheme
that is used in CutMix-Seg.

When comparing the results of our approach
w/o and w/ CutMix augmentation,
we have the following observation:
the CutMix augmentation is more important
for the scenario with fewer labeled data.
For example, with ResNet-$50$,
the gain $3.77\%$ under the $1/16$ partition
is higher than $0.47\%$ under the $1/8$ partition.

\emph{Cityscapes}:
Table~\ref{tab:sota-City-subset} illustrates the comparison results on the
Cityscapes \texttt{val} set. 
We do not have the results for CutMix-Seg
as the official CutMix-Seg implementation only supports single-GPU training
and it is not feasible to run CutMix-Seg with DeepLabv$3$+
on Cityscapes due to the GPU memory limit.
In comparison to other SOTA methods,
our method achieves the best performance among all partition protocols with both ResNet-$50$ and ResNet-$101$ backbones.
For example, our method w/ CutMix augmentation obtains $80.08$\% under the $1/2$ partition with ResNet-$101$ backbone, which outperforms GCT by $1.50\%$.
We report the additional results on HRNet in Table~\ref{tab:sota-HRNet-subset}.

\begin{figure}
\centering
\footnotesize
\begin{tikzpicture}[scale=1]
\pgfplotsset{compat=1.4}
\begin{axis}[
    ybar,
    height=5cm,
    width=8cm,
    enlargelimits=0.0,
    enlarge x limits=0.5,
    bar width=18pt,
    ylabel={mIoU ($\%$)},
    ymin=80, ymax=83,
    symbolic x coords={DeepLabv3+, HRNet-W48},
    xtick=data,
    nodes near coords,
    nodes near coords align={vertical},
    legend cell align={left},
    legend columns=-1,
    y label style={at={(-0.08,0.5)}},
    legend image code/.code={
      \draw[#1] (0cm,-0.1cm) rectangle (0.4cm,0.15cm);
    },
    legend style={at={(0.05,0.75)},anchor=south west, font=\footnotesize},
    xticklabel style={rotate=0,font=\footnotesize, align=center},
]
\definecolor{bananamania}{rgb}{0.98, 0.91, 0.71}
\definecolor{mossgreen}{rgb}{0.68, 0.87, 0.68}
\definecolor{mintgreen}{rgb}{0.6, 1.0, 0.6}
\definecolor{palegreen}{rgb}{0.6, 0.98, 0.6}
\definecolor{bananayellow}{rgb}{1.0, 0.88, 0.21}
\definecolor{bluebell}{rgb}{0.64, 0.64, 0.82}
\definecolor{carnationpink}{rgb}{1.0, 0.65, 0.79}

\addplot +[bar shift=-0.42cm] coordinates {(DeepLabv3+, 80.40) (HRNet-W48,80.65)};
\addplot +[bar shift=+0.42cm] coordinates {(DeepLabv3+, 81.54) (HRNet-W48,82.41)};

\addlegendentry{\, Baseline}    
\addlegendentry{\, Ours}
\end{axis}
\end{tikzpicture}

\vspace{-2mm}
\captionsetup{font={small}}
  \caption{\textbf{Improving the fully-supervised baselines.}
  The baseline models (DeepLabv$3$+ with ResNet-$101$
  and HRNet-W$48$)
  are trained using the full Cityscapes \texttt{train} set.
  Our approach uses
  $\sim3,000$ images from Cityscapes \texttt{coarse} set 
  as an additional unlabeled set 
  for training.
  The superiority of our approach
  implies that our approach works well
  on the relatively large labeled data.}
 \vspace{-2mm}
\label{fig:sota-city-fullset}
\end{figure}
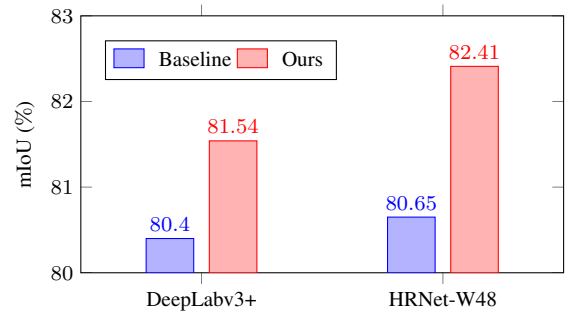

\subsection{Improving Full- and Few-Supervision}

\noindent\textbf{Full-supervision.}
We verify our method using the full Cityscapes \texttt{train} set ($\sim2,975$ images) and randomly sample $3,000$ images from the  Cityscapes \texttt{coarse} set as the unlabeled set.
For the unlabeled set, we do not use their coarsely annotated ground truth.
Figure~\ref{fig:sota-city-fullset} illustrates the results on the
Cityscapes \texttt{val} set with single-scale evaluation.
We can see that even
with a large amount of labeled data,
our approach could still benefit from training with unlabeled data,
and our approach also works well
on the state-of-the-art segmentation network HRNet.

\renewcommand{\arraystretch}{1.4}
\begin{table*}[]
\centering\setlength{\tabcolsep}{15pt}
\centering
\captionsetup{font={small}}
\caption{\textbf{Ablation study of different loss combinations}
on PASCAL VOC $2012$ and Cityscapes.
The results are obtained under
the 1/8
data partition protocol
and the observations are consistent for
other partition protocols.
$\mathcal{L}_s$ represents the supervision loss on the labeled set.
$\mathcal{L}_{cps}^l$ ($\mathcal{L}_{cps}^u$) represents the cross pseudo supervision loss on the labeled (unlabeled) set.
$\mathcal{L}_{cpc}^l$ ($\mathcal{L}_{cpc}^u$) represents the cross probability consistency loss on the labeled (unlabeled) set.
The overall performance with the cross pseudo supervision loss on both the labeled and unlabeled data
is the best.
}
\vspace{-2mm}
\label{tab:abla-of-losses}
\footnotesize
\begin{tabular}{c|c|c|c|c|c|c|c|c}
\shline 
\multicolumn{5}{c|}{Losses}  & \multicolumn{2}{c|}{PASCAL VOC $2012$} & \multicolumn{2}{c}{Cityscapes} \\ \hline
$\mathcal{L}_s$ & $\mathcal{L}_{cps}^l$ & $\mathcal{L}_{cps}^u$ & $\mathcal{L}_{cpc}^l$ & $\mathcal{L}_{cpc}^u$ &  \multicolumn{1}{c|}{ResNet-$50$} & \multicolumn{1}{c|}{ResNet-$101$} & \multicolumn{1}{c|}{ResNet-$50$} & \multicolumn{1}{c}{ResNet-$101$}  \\ \shline
\cmark &  &  &  &  &  $69.43$ & $72.21$ & $70.32$  & $72.19$\\
\cmark & \cmark &  &  &  &  $69.99$ & $72.98$ &  $71.73$  & $73.08$\ \\
\cmark &  & \cmark &  &  &  $73.00$ & $75.83$ & $73.97$ & $75.28$  \\
\cmark & \cmark & \cmark &  &  & $\bf{73.20}$ & $\bf{75.85}$ & $\bf{74.39}$ & $\bf{75.71}$ \\
\cmark &  &  & \cmark & \cmark &  $71.23$ & $74.01$ & $72.03$  & $73.77$  \\ \hline
\end{tabular}
\end{table*}

\renewcommand{\arraystretch}{1.4}
\begin{table}[]
\centering\setlength{\tabcolsep}{7pt}
\captionsetup{font={small}}
\caption{\textbf{Comparison for few-supervision} on PASCAL VOC $2012$.
We follow the same partition protocols 
provided in PseudoSeg~\cite{zou2020pseudoseg}.
The results of all the other methods are from~\cite{zou2020pseudoseg}.
\vspace{-4mm}
}
\footnotesize
\begin{tabular}{l|cccc}
\shline
\multirow{2}{*}{Method} & \multicolumn{4}{c}{$\#$(labeled samples)} \\ \cline{2-5}
& 732 & 366 & 183 & 92  \\
\shline
AdvSemSeg~\cite{hung2018adversarial} & $65.27$ & $59.97$ & $47.58$ & $39.69$ \\
CCT~\cite{ouali2020semi} & $62.10$ & $58.80$ & $47.60$ & $33.10$ \\
MT~\cite{tarvainen2017mean} & $69.16$ & $63.01$ & $55.81$ & $48.70$ \\
GCT~\cite{ke2020guided} & $70.67$ & $64.71$ & $54.98$ & $46.04$ \\
VAT~\cite{VAT} & $63.34$ & $56.88$ & $49.35$ & 36.92 \\
CutMix-Seg~\cite{french2020semi} & $69.84$ & $68.36$ & $63.20$ & $55.58$ \\ 
PseudoSeg~\cite{zou2020pseudoseg} & $72.41$ & $69.14$ & $65.50$ & $57.60$ \\ \hline
Ours (w/ CutMix  Aug.) & $\mathbf{75.88}$ & $\mathbf{71.71}$ & $\mathbf{67.42}$ &  $\mathbf{64.07}$\\
\hline
\end{tabular}
\label{tab:voc_sota_ours_unlabeled}
\end{table}

\noindent\textbf{Few-supervision.}
We study the performance of our method on PASCAL VOC $2012$ with very few supervision
by following the same partition protocols adopted in PseudoSeg~\cite{zou2020pseudoseg}.
PseudoSeg randomly samples $1/2$, $1/4$, $1/8$, and $1/16$ of images in the standard training set (around $1.5$k images) to construct the labeled set. The remaining images in the standard training set, together with the images in the augmented set~\cite{hariharan2011sbd} (around $9$k images), are
used as the unlabeled set.

We only report the results of our approach w/ CutMix augmentation
as CutMix is important for few supervision.
Results are listed in Table~\ref{tab:voc_sota_ours_unlabeled},
where all methods use ResNet-$101$ as the backbone except CCT that uses ResNet-$50$.
We can see that our approach performs the best and is superior to CutMix-Seg again
on the few labeled case.
Our approach is also better than PseudoSeg
that uses a complicated scheme to compute
the pseudo segmentation map.
We believe that the reason
comes from 
that our approach uses network perturbation
and cross pseudo supervision
while PseudoSeg uses a single network 
with input perturbation.

\subsection{Empirical Study}

\begin{figure}
\centering
\footnotesize
\begin{tikzpicture}[scale=1]
\begin{axis}[
footnotesize,
legend columns=1, legend style={at={(0.66,0.4)}, nodes={scale=0.8},
anchor=north}, ymin=68,
y label style={at={(0.16,0.5)}},
    height=3.8cm,
    width=5cm,
xtick={0, 0.5, 1.0, 1.5, 2.0}, 
xticklabels={0, 0.5, 1.0, 1.5, 2.0},
ymajorgrids=true]
\addplot+[sharp plot, line width=1pt, mark size=1pt]
table
{
X Y
0.0 69.43
0.5 72.07
1.0 72.89
1.5 73.20
2.0 72.85
};

\addplot+[sharp plot, line width=1pt, mark size=1pt, color=red, mark=diamond]
table
{
X Y
0.0 72.21
0.5 75.53
1.0 75.46
1.5 75.83
2.0 75.65
};

\addlegendentry{ResNet-50}    
\addlegendentry{ResNet-101} 
\end{axis}
\end{tikzpicture}
~~~~
\begin{tikzpicture}[scale=1]
\begin{axis}[legend columns=1, legend style={at={(0.66,0.4)}, 
nodes={scale=0.8}, 
anchor=north}, ymin=69,
y label style={at={(0.16,0.5)}},
    height=3.8cm,
    width=5cm,
xtick={0, 1, 2, 3, 4, 5, 6}, 
xticklabels={0, 1, 2, 3, 4, 5, 6},
ymajorgrids=true]
\addplot+[sharp plot, line width=1pt, mark size=1pt]
table
{
X Y
0.0 70.32
1.0 72.17
3.0 73.05
5.0 73.69
6.0 74.39
};
\addplot+[sharp plot, line width=1pt, mark size=1pt, color=red, mark=diamond]
table
{
X Y
0.0 72.19
1.0 74.25
3.0 75.15
5.0 75.62
6.0 75.71
};
\addlegendentry{ResNet-50}    
\addlegendentry{ResNet-101} 
\end{axis}
\end{tikzpicture}
\vspace{-2mm}
\captionsetup{font={small}}
  \caption{\textbf{Illustration
  on how the trade-off weight $\lambda$
  (x-axis)
  affects the mIoU score (y-axis)} on
  PASCAL VOC $2012$ (left) and 
  Cityscapes (right). All results are evaluated under the
  $1/8$ partition protocol.}
  \label{fig:different-lambda}
  \vspace{-6mm}
\end{figure}
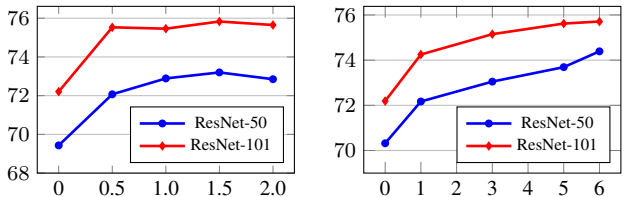

\noindent \textbf{Cross pseudo supervision.}
We investigate the influence of applying the proposed cross pseudo supervision loss to labeled set ($\mathcal{L}_{cps}^{l}$) or unlabeled set ($\mathcal{L}_{cps}^{u}$)
in the Table~\ref{tab:abla-of-losses}.
We can see that cross pseudo supervision loss on the unlabeled set brings more significant
improvements than cross pseudo supervision loss on the labeled set in most cases.
For example, with ResNet-$50$,
cross pseudo supervision loss on the labeled set improves the performance of the baseline by $0.56\%$ ($1.41\%$) while cross pseudo supervision loss on the unlabeled set improves by $3.57\%$ ($4.07\%$) on PASCAL VOC $2012$ (Cityscapes).
The performance with cross pseudo supervision loss on both labeled set and unlabeled set
is overall the best.

\vspace{0.1cm}
\noindent \textbf{Comparison with cross probability consistency.}
We compare our method with the cross probability consistency
on the last $2$ rows of Table~\ref{tab:abla-of-losses}.
We can see that our cross pseudo supervision outperforms the
cross probability consistency on both benchmarks.
For example, on Cityscapes, cross pseudo supervision outperforms cross probability consistency by $2.36\%$ ($1.94\%$) when applied to both labeled and unlabeled sets with ResNet-$50$ (ResNet-$101$).

\vspace{0.1cm}
\noindent \textbf{The trade-off weight $\lambda$.}
We investigate the influence of different $\lambda$ that is
used to balance the supervision loss and cross pseudo supervision loss as shown in Equation~\ref{eq:whole-loss}.
From Figure~\ref{fig:different-lambda},
we can see that $\lambda$ = $1.5$ performs best on PASCAL VOC $2012$
and $\lambda$ = $6$ performs best on Cityscapes.
We use $\lambda$ = $1.5$ and $\lambda$ = $6$
in our approach
for all the experiments.

\begin{table}[t]
\centering
\footnotesize
\setlength{\tabcolsep}{0.2cm}
\captionsetup{font={small}}
\caption{\textbf{Comparison with
single-network pseudo supervision} on PASCAL VOC $2012$ \texttt{val}.
SPS = single-network pseudo supervision.
All methods are based on DeepLabv3+ are with ResNet-$50$.
We can see that for both the two cases,
w/ and w/o CutMix augmentation,
our approach outperforms
the single-network pseudo supervision.}
\vspace{-2mm}
\label{tab:abla-single-network-sup}
\begin{tabular}{l|c|c|c|c} 
\shline
Method & 1/16 & 1/8 & 1/4 & 1/2 \\
\shline
SPS (w/o CutMix Aug.) & $ 59.54 $ & $ 69.05 $ & $ 72.55 $ & $ 75.17 $ \\
Ours (w/o CutMix Aug.) & $\mathbf{68.21}$ & $\bf{73.20}$ & $\bf{74.24}$ & $\bf{75.91}$\\\hline
SPS (w/ CutMix Aug.) & ${65.62}$ & ${71.27} $ & ${73.70} $ & ${74.87}$  \\
Ours (w/ CutMix Aug.) & $\bf{71.98}$ & $\bf{73.67} $ & $\bf{74.90} $ & $\bf{76.15}$ \\
\hline
\end{tabular}
\end{table}

\vspace{0.1cm}
\noindent \textbf{Single-network pseudo supervision vs. cross pseudo supervision}.
We compare the proposed approach with single-network pseudo supervision in Table~\ref{tab:abla-single-network-sup}.
We can see that our method outperforms the single-network pseudo supervision scheme
either with CutMix augmentation or not.
The single-network pseudo supervision with the CutMix augmentation
is similar to 
the application of FixMatch~\cite{SohnBLZCCKZR20}
to semantic segmentation
(as done in PseudoSeg).
We think that this is one of the main reason
that our approach is superior
to PseudoSeg.

\renewcommand{\arraystretch}{1.4}
\begin{table}[]
\centering\setlength{\tabcolsep}{7pt}
\captionsetup{font={small}}
\caption{\textbf{Combination with self-training.}
The CutMix augmentation is not used.
We can see that the combination 
gets improves over both self-training and our approach. }
\vspace{-2mm}
\label{tab:abla-self-training}
\footnotesize
\begin{tabular}{l|c|c|c|c}
\shline
\multirow{2}{*}{Method} & \multicolumn{2}{c|}{ResNet-$50$} & \multicolumn{2}{c}{ResNet-$101$} \\\cline{2-5}
& $1/4$  & $1/2$  & $1/4$  & $1/2$  \\
\shline
\multicolumn{5}{l}{\emph{PASCAL VOC $2012$}} \\\hline
Ours & $74.24$  & $75.91$  & $77.55$ & $78.64$  \\ 
Self-Training  & $74.47$ & $75.97$  & $76.63$ & $78.15$ \\ 
Ours + Self-Training  & $\bf{74.96}$ & $\bf{76.60}$ & $\bf{77.60}$ & $\bf{78.76}$ \\ 
\hline
\multicolumn{5}{l}{\emph{Cityscapes}} \\\hline
Ours & $76.85$ & $78.64$  & $77.41$ & $80.08$ \\ 
Self-Training  & $75.88$ & $77.64$  & $77.55$ & $79.46$ \\ 
Ours + Self-Training  & $\bf{77.40}$ & $\bf{79.25}$ & $\bf{79.16}$  & $\bf{80.17}$\\ 
\hline
\end{tabular}
\vspace{-2mm}
\end{table}

\vspace{0.1cm}
\noindent \textbf{Combination/comparison with self-training.}
We empirically study the combination of our method and the conventional self-training~\cite{xie2020self}.
Results on both benchmarks are summarized in Table~\ref{tab:abla-self-training}. 
We can see that the combination of self-training and our approach
outperforms both our method only and self-training only.
The superiority implies that
our approach is complementary to self-training.

As the self-training scheme consists of multiple stages (train over labeled set $\rightarrow$ predict pseudo labels for unlabeled set $\rightarrow$ retrain over labeled and unlabeled set with pseudo labels),
it takes more training epochs
than our approach.
For a fairer comparison with self-training,
we train our method for more epochs (denoted as Ours$^+$)
to ensure our training epochs are also comparable with self-training.
According to the results shown in Figure~\ref{fig:bar-compare-self-training},
we can see that ours$^+$ consistently outperforms self-training under various partition protocols.
We guess that the reason lies in the consistency regularization in our approach.

\newcommand\distance{0.005}
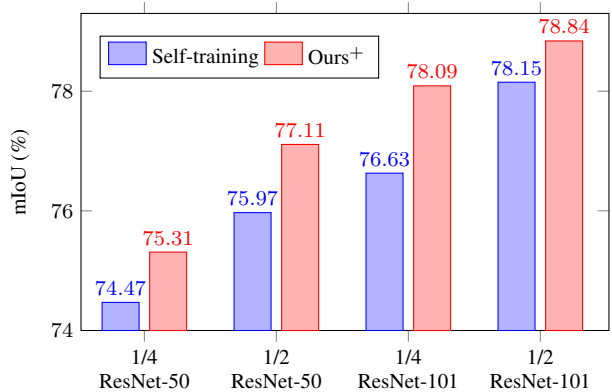
\begin{figure}
\centering
\begin{tikzpicture}[scale=1]                 
\footnotesize
\begin{axis}[
legend columns=-1,
height=5.8cm,
width=8.6cm,
legend style={at={(0.3,0.93)},
anchor=north,font=\footnotesize},
legend cell align={left},
y label style={at={(0.06,0.5)}},
ylabel={mIoU (\%)},
xtick={0.12, 0.37, 0.62, 0.87},
xticklabels={1/4 \\ResNet-50, 1/2 \\ ResNet-50, 1/4 \\ ResNet-101, 1/2 \\ ResNet-101},
xticklabel style={rotate=0,font=\footnotesize, align=center},
xmin=0,
xmax=1,
ymin=74,
ymax=79.3,
bar width=14pt,
ybar,
nodes near coords,
legend entries={Self-training, Ours$^+$},
legend image code/.code={\draw[#1] (0cm,-0.1cm) rectangle (0.5cm,0.2cm);}
]

\addplot
coordinates
{
(0.12-\distance, 74.47)
(0.37-\distance, 75.97)
(0.62-\distance, 76.63)
(0.87-\distance, 78.15)
};

\addplot
coordinates
{
(0.12+\distance, 75.31)
(0.37+\distance, 77.11)
(0.62+\distance, 78.09)
(0.87+\distance, 78.84)
};
  
\end{axis}
\end{tikzpicture}
\vspace{-2mm}
\captionsetup{font={small}}
  \caption{\textbf{Comparison with self-training}
  on PASCAL VOC $2012$. 
  The self-training approach
  is a two-stage approach
  which takes more training epochs.
  For a fair comparison,
  we train our approach with more training epochs
  (denoted by `Ours$^+$) 
  so that their epochs are comparable.
  The CutMix augmentation is not used.}
\label{fig:bar-compare-self-training}
\end{figure} 

\begin{figure}[htp]
\centering
\includegraphics[width=\linewidth]{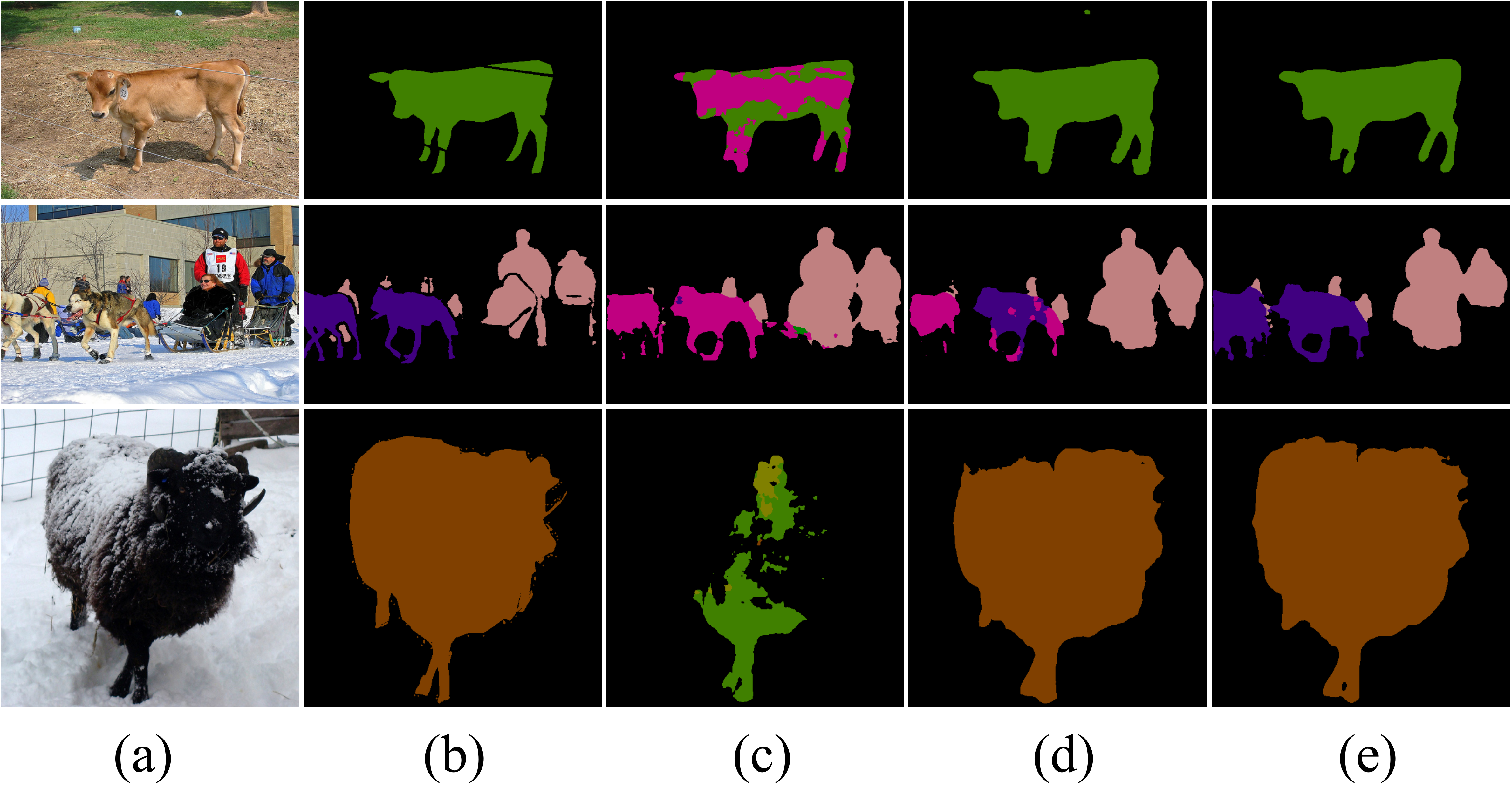}
\vspace{-4mm}
\captionsetup{font={small}}
\caption{\textbf{Example qualitative results from PASCAL VOC} $2012$. (a) input, (b) ground truth, (c) supervised only, (d) ours (w/o CutMix Aug.), and (e) ours (w/ CutMix Aug.).
All the approaches use DeepLabv$3$+ with ResNet-$101$
as the segmentation network. }
\label{fig:vocvis}
\vspace{-3mm}
\end{figure}

\subsection{Qualitative Results}
Figure~\ref{fig:vocvis} visualizes some segmentation results on PASCAL VOC $2012$.
We can see the supervised baseline, shown in the Figure~\ref{fig:vocvis} column (c), mis-classifies many pixels due to limited labeled training samples.
For example, in the $1$-st row, the supervised only method (column (c)) mistakenly classifies many cow pixels as horse pixels while our method w/o CutMix augmentation (column (d))
fixes these errors.
In the $2$-nd row, both the supervised baseline and our method w/o CutMix augmentation, mislabel some dog pixels as horse pixels while
our method w/ CutMix augmentation (column (e)) successfully corrects these errors.

\section{Conclusion}
We present a simple but effective 
semi-supervised segmentation approach, cross pseudo supervision.
Our approach imposes the consistency
between two networks
with the same structure
and different initialization, 
by using the one-hot pseudo segmentation map
obtained from one network 
to supervise the other network.
On the other hand,
the unlabeled data with pseudo segmentation map,
which is more accurate in the later training stage,
serves as expanding the training data
to improve the performance. \\[0.8mm]
\noindent \textbf{Acknowledgments:} This work is supported by the National Key Research and Development Program of China (2017YFB1002601, 2016QY02D0304), National Natural Science Foundation of China (61375022, 61403005, 61632003), Beijing Advanced Innovation Center for Intelligent Robots and Systems (2018IRS11), and PEK-SenseTime Joint Laboratory of Machine Vision.

\newpage

\section*{Appendix}
\appendix
\section{More Implementation Details}
\noindent \textbf{Training details.} 
The crop size for PASCAL VOC $2012$ and Cityscapes are $512 \times 512$ and $800 \times 800$, respectively. For the multi-scale data augmentation, we randomly select scale from \{$0.5$, $0.75$, $1$, $1.25$, $1.5$, $1.75$\}. For Cityscapes dataset, we use OHEM loss as the supervision loss ($\mathcal{L}_s$), and cross entropy loss as the cross pseudo supervision loss ($\mathcal{L}_{cps}$). 

\vspace{1mm}
\noindent \textbf{Training strategy.} 
We use
the similar training strategy
as GCT~\cite{ke2020guided} for semi-supervised 
segmentation. 
In the supervised baseline
for all the partition protocols, 
we use the batch size $8$.
We ensure that the iteration number
is the same as semi-supervised methods\footnote{In GCT~\cite{ke2020guided},
the supervised baseline 
uses the batch size $16$,
and the number of iterations
is much smaller than half of the number of iterations
in the semi-supervised methods.
Therefore, their supervised baseline results
are worse than ours
(we sure the same number of iterations and each iteration has the same number, $8$,
of labeled samples).}.
For semi-supervised methods, 
at each iteration,
we sample additional $8$ unlabeled samples.
Our method
and all the other semi-supervised methods in Table 1 and Table 2 of the main paper
follow the same training strategy.

\section{Network Perturbation}

Our cross pseudo supervision approach (CPS) 
includes two perturbed segmentation networks, $f(\boldsymbol{\uptheta}_1)$ and $f(\boldsymbol{\uptheta}_2)$, which 
are of the same architecture
and initialized differently.
In the main paper,
we pointed out that
the pseudo segmentation results 
from the two networks
are perturbed.

\begin{figure}[t]
  \centering
  \includegraphics[width=0.9\linewidth]{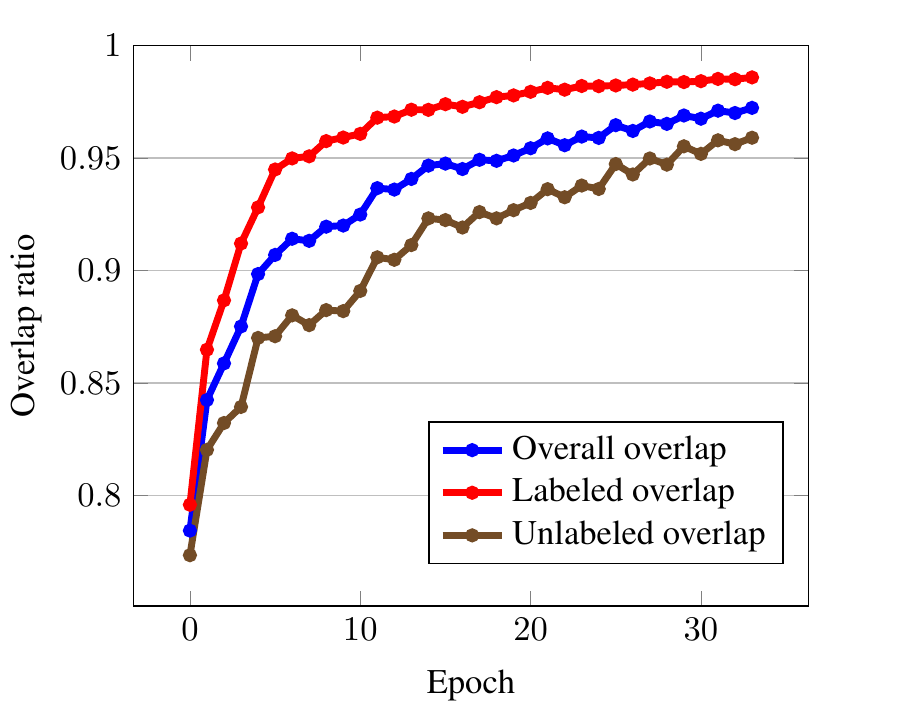}
  \captionsetup{font={small}}
  \caption{\textbf{Prediction overlap of the two networks on PASCAL VOC $2012$ under the $1/8$ partition}. We use DeepLabv$3$+ with ResNet-$50$ as the segmentation network. 
  We only calculate the overlap ratio in the object region, 
  and the pixels belong to the `background' class are ignored.}
  \label{fig:supp-overlap}
  \end{figure}

We empirically
show 
the perturbation
using the overlap ratio between them
during training.
The overlap ratio on the labeled set, the unlabeled set and the whole set are given in Figure~\ref{fig:supp-overlap}. 
We can see that 
(1) the overlap ratio is small at the early training stage and (2) increases during the later training stage.
The small overlap ratio at the early stage 
helps avoid the case the segmentation network converges towards 
a wrong direction.
The large overlap ratio at the later stage
implies that the pseudo segmentation results of the two segmentation networks
are more accurate. 

{\small
\bibliographystyle{ieee_fullname}
\bibliography{semiseg}

\begin{thebibliography}{10}\itemsep=-1pt

\bibitem{Agrawala70}
Ashok~K. Agrawala.
\newblock Learning with a probabilistic teacher.
\newblock {\em {IEEE} Trans. Inf. Theory}, 16(4):373--379, 1970.

\bibitem{BerthelotCGPOR19}
David Berthelot, Nicholas Carlini, Ian Goodfellow, Nicolas Papernot, Avital
  Oliver, and Colin~A Raffel.
\newblock Mixmatch: A holistic approach to semi-supervised learning.
\newblock In {\em NIPS}, pages 5049--5059, 2019.

\bibitem{CSZ2006}
Olivier Chapelle, Bernhard Sch{\"{o}}lkopf, and Alexander Zien, editors.
\newblock {\em Semi-Supervised Learning}.
\newblock The {MIT} Press, 2006.

\bibitem{chen2018deeplab}
Liang-Chieh Chen, George Papandreou, Iasonas Kokkinos, Kevin Murphy, and Alan~L
  Yuille.
\newblock Deeplab: Semantic image segmentation with deep convolutional nets,
  atrous convolution, and fully connected crfs.
\newblock {\em PAMI}, 2018.

\bibitem{chen2018encoder}
Liang-Chieh Chen, Yukun Zhu, George Papandreou, Florian Schroff, and Hartwig
  Adam.
\newblock Encoder-decoder with atrous separable convolution for semantic image
  segmentation.
\newblock In {\em ECCV}, 2018.

\bibitem{ChenLCCCZAS20}
Maxwell~D Collins, Ekin~D Cubuk, Hartwig~Adam Barret~Zoph, and Jonathon Shlens.
\newblock Naive-student: Leveraging semi-supervised learning in video sequences
  for urban scene segmentation.
\newblock In {\em ECCV}. Springer, 2020.

\bibitem{cordts2016cityscapes}
Marius Cordts, Mohamed Omran, Sebastian Ramos, Timo Rehfeld, Markus Enzweiler,
  Rodrigo Benenson, Uwe Franke, Stefan Roth, and Bernt Schiele.
\newblock The cityscapes dataset for semantic urban scene understanding.
\newblock In {\em CVPR}, 2016.

\bibitem{everingham2015pascal}
Mark Everingham, SM~Ali Eslami, Luc Van~Gool, Christopher~KI Williams, John
  Winn, and Andrew Zisserman.
\newblock The pascal visual object classes challenge: A retrospective.
\newblock {\em IJCV}, 111(1):98--136, 2015.

\bibitem{FengZCTSM20}
Zhengyang Feng, Qianyu Zhou, Guangliang Cheng, Xin Tan, Jianping Shi, and
  Lizhuang Ma.
\newblock Semi-supervised semantic segmentation via dynamic self-training and
  class-balanced curriculum.
\newblock {\em CoRR}, abs/2004.08514, 2020.

\bibitem{Fralick67}
Stanley~C. Fralick.
\newblock Learning to recognize patterns without a teacher.
\newblock {\em {IEEE} Trans. Inf. Theory}, 13(1):57--64, 1967.

\bibitem{french2020semi}
Geoff French, Timo Aila, Samuli Laine, Michal Mackiewicz, and Graham Finlayson.
\newblock Semi-supervised semantic segmentation needs strong, high-dimensional
  perturbations.
\newblock In {\em BMVC}, 2020.

\bibitem{hariharan2011sbd}
Bharath Hariharan, Pablo Arbel{\'a}ez, Lubomir Bourdev, Subhransu Maji, and
  Jitendra Malik.
\newblock Semantic contours from inverse detectors.
\newblock In {\em 2011 International Conference on Computer Vision}, pages
  991--998. IEEE, 2011.

\bibitem{hung2018adversarial}
Wei-Chih Hung, Yi-Hsuan Tsai, Yan-Ting Liou, Yen-Yu Lin, and Ming-Hsuan Yang.
\newblock Adversarial learning for semi-supervised semantic segmentation.
\newblock In {\em BMVC}, 2018.

\bibitem{ibrahim2020semi}
Mostafa~S Ibrahim, Arash Vahdat, Mani Ranjbar, and William~G Macready.
\newblock Semi-supervised semantic image segmentation with self-correcting
  networks.
\newblock In {\em CVPR}, pages 12715--12725, 2020.

\bibitem{Scudder65a}
H.~J.~Scudder III.
\newblock Probability of error of some adaptive pattern-recognition machines.
\newblock {\em {IEEE} Trans. Inf. Theory}, 11(3):363--371, 1965.

\bibitem{ioffe2015batch}
Sergey Ioffe and Christian Szegedy.
\newblock Batch normalization: Accelerating deep network training by reducing
  internal covariate shift.
\newblock {\em arXiv preprint arXiv:1502.03167}, 2015.

\bibitem{ke2020guided}
Zhanghan Ke, Di Qiu, Kaican Li, Qiong Yan, and Rynson~WH Lau.
\newblock Guided collaborative training for pixel-wise semi-supervised
  learning.
\newblock In {\em ECCV}, 2020.

\bibitem{ke2019dual}
Zhanghan Ke, Daoye Wang, Qiong Yan, Jimmy Ren, and Rynson~WH Lau.
\newblock Dual student: Breaking the limits of the teacher in semi-supervised
  learning.
\newblock In {\em ICCV}, pages 6728--6736, 2019.

\bibitem{KimJP20}
Jongmok Kim, Jooyoung Jang, and Hyunwoo Park.
\newblock Structured consistency loss for semi-supervised semantic
  segmentation.
\newblock {\em CoRR}, abs/2001.04647, 2020.

\bibitem{kirillov2020pointrend}
Alexander Kirillov, Yuxin Wu, Kaiming He, and Ross Girshick.
\newblock Pointrend: Image segmentation as rendering.
\newblock In {\em CVPR}, pages 9799--9808, 2020.

\bibitem{LaineA17}
Samuli Laine and Timo Aila.
\newblock Temporal ensembling for semi-supervised learning.
\newblock {\em ICLR}, 2017.

\bibitem{lee2013pseudo}
Dong-Hyun Lee.
\newblock Pseudo-label: The simple and efficient semi-supervised learning
  method for deep neural networks.
\newblock In {\em ICMLW}, 2013.

\bibitem{long2015fully}
Jonathan Long, Evan Shelhamer, and Trevor Darrell.
\newblock Fully convolutional networks for semantic segmentation.
\newblock In {\em CVPR}, 2015.

\bibitem{mendel2020semi}
Robert Mendel, Luis~Antonio de Souza, David Rauber, Jo{\~a}o~Paulo Papa, and
  Christoph Palm.
\newblock Semi-supervised segmentation based on error-correcting supervision.
\newblock In {\em ECCV}, pages 141--157. Springer, 2020.

\bibitem{MittalIB19}
Sudhanshu Mittal, Maxim Tatarchenko, and Thomas Brox.
\newblock Semi-supervised semantic segmentation with high- and low-level
  consistency.
\newblock {\em CoRR}, abs/1908.05724, 2019.

\bibitem{VAT}
Takeru Miyato, Shin-ichi Maeda, Shin Ishii, and Masanori Koyama.
\newblock Virtual adversarial training: a regularization method for supervised
  and semi-supervised learning.
\newblock {\em TPAMI}, 2018.

\bibitem{ouali2020semi}
Yassine Ouali, C{\'e}line Hudelot, and Myriam Tami.
\newblock Semi-supervised semantic segmentation with cross-consistency
  training.
\newblock In {\em CVPR}, pages 12674--12684, 2020.

\bibitem{SohnBLZCCKZR20}
Kihyuk Sohn, David Berthelot, Chun{-}Liang Li, Zizhao Zhang, Nicholas Carlini,
  Ekin~D. Cubuk, Alex Kurakin, Han Zhang, and Colin Raffel.
\newblock Fixmatch: Simplifying semi-supervised learning with consistency and
  confidence.
\newblock {\em CoRR}, abs/2001.07685, 2020.

\bibitem{souly2017semi}
Nasim Souly, Concetto Spampinato, and Mubarak Shah.
\newblock Semi supervised semantic segmentation using generative adversarial
  network.
\newblock In {\em ICCV}, pages 5688--5696, 2017.

\bibitem{sun2019deep}
Ke Sun, Bin Xiao, Dong Liu, and Jingdong Wang.
\newblock Deep high-resolution representation learning for human pose
  estimation.
\newblock In {\em CVPR}, pages 5693--5703, 2019.

\bibitem{takikawa2019gated}
Towaki Takikawa, David Acuna, Varun Jampani, and Sanja Fidler.
\newblock Gated-scnn: Gated shape cnns for semantic segmentation.
\newblock In {\em ICCV}, pages 5229--5238, 2019.

\bibitem{tarvainen2017mean}
Antti Tarvainen and Harri Valpola.
\newblock Mean teachers are better role models: Weight-averaged consistency
  targets improve semi-supervised deep learning results.
\newblock In {\em NIPS}, 2017.

\bibitem{vaswani2017attention}
Ashish Vaswani, Noam Shazeer, Niki Parmar, Jakob Uszkoreit, Llion Jones,
  Aidan~N Gomez, {\L}ukasz Kaiser, and Illia Polosukhin.
\newblock Attention is all you need.
\newblock In {\em NIPS}, 2017.

\bibitem{wang2020deep}
Jingdong Wang, Ke Sun, Tianheng Cheng, Borui Jiang, Chaorui Deng, Yang Zhao,
  Dong Liu, Yadong Mu, Mingkui Tan, Xinggang Wang, et~al.
\newblock Deep high-resolution representation learning for visual recognition.
\newblock {\em TPAMI}, 2020.

\bibitem{xie2020self}
Qizhe Xie, Minh-Thang Luong, Eduard Hovy, and Quoc~V Le.
\newblock Self-training with noisy student improves imagenet classification.
\newblock In {\em CVPR}, pages 10687--10698, 2020.

\bibitem{yu2015multi}
Fisher Yu and Vladlen Koltun.
\newblock Multi-scale context aggregation by dilated convolutions.
\newblock {\em ICLR}, 2016.

\bibitem{yuan2019object}
Yuhui Yuan, Xilin Chen, and Jingdong Wang.
\newblock Object-contextual representations for semantic segmentation.
\newblock {\em arXiv preprint arXiv:1909.11065}, 2019.

\bibitem{yuan2018ocnet}
Yuhui Yuan and Jingdong Wang.
\newblock Ocnet: Object context network for scene parsing.
\newblock {\em arXiv:1809.00916}, 2018.

\bibitem{yuan2020segfix}
Yuhui Yuan, Jingyi Xie, Xilin Chen, and Jingdong Wang.
\newblock Segfix: Model-agnostic boundary refinement for segmentation.
\newblock In {\em ECCV}, pages 489--506. Springer, 2020.

\bibitem{yun2019cutmix}
Sangdoo Yun, Dongyoon Han, Seong~Joon Oh, Sanghyuk Chun, Junsuk Choe, and
  Youngjoon Yoo.
\newblock Cutmix: Regularization strategy to train strong classifiers with
  localizable features.
\newblock In {\em ICCV}, pages 6023--6032, 2019.

\bibitem{zhao2017pyramid}
Hengshuang Zhao, Jianping Shi, Xiaojuan Qi, Xiaogang Wang, and Jiaya Jia.
\newblock Pyramid scene parsing network.
\newblock In {\em CVPR}, 2017.

\bibitem{ZhuZWZHZMLS20}
Yi Zhu, Zhongyue Zhang, Chongruo Wu, Zhi Zhang, Tong He, Hang Zhang, R.
  Manmatha, Mu Li, and Alexander~J. Smola.
\newblock Improving semantic segmentation via self-training.
\newblock {\em CoRR}, abs/2004.14960, 2020.

\bibitem{ZophGLCLCL20}
Barret Zoph, Golnaz Ghiasi, Tsung{-}Yi Lin, Yin Cui, Hanxiao Liu, Ekin~D.
  Cubuk, and Quoc~V. Le.
\newblock Rethinking pre-training and self-training.
\newblock {\em CoRR}, abs/2006.06882, 2020.

\bibitem{zou2020pseudoseg}
Yuliang Zou, Zizhao Zhang, Han Zhang, Chun-Liang Li, Xiao Bian, Jia-Bin Huang,
  and Tomas Pfister.
\newblock Pseudoseg: Designing pseudo labels for semantic segmentation.
\newblock {\em arXiv preprint arXiv:2010.09713}, 2020.

\end{thebibliography}
}

\end{document}